\documentclass{article} 
\usepackage{tencent_ailab_tech_report}
\usepackage[colorlinks = true,
            linkcolor = blue,
            urlcolor  = blue,
            citecolor = blue,
            anchorcolor = blue]{hyperref}

\usepackage{graphicx}
\usepackage{caption}
\usepackage{subcaption}
\usepackage{booktabs} 
\usepackage{xurl}

\usepackage{listings}
\usepackage{xcolor}

\lstdefinelanguage{json}{
    basicstyle=\ttfamily\small,
    numbers=left,
    numberstyle=\tiny,
    stepnumber=1,
    numbersep=5pt,
    showstringspaces=false,
    breaklines=true,
    frame=lines,
    backgroundcolor=\color{gray!10},
    literate=
     *{0}{{{\color{blue}0}}}{1}
      {1}{{{\color{blue}1}}}{1}
      {2}{{{\color{blue}2}}}{1}
      {3}{{{\color{blue}3}}}{1}
      {4}{{{\color{blue}4}}}{1}
      {5}{{{\color{blue}5}}}{1}
      {6}{{{\color{blue}6}}}{1}
      {7}{{{\color{blue}7}}}{1}
      {8}{{{\color{blue}8}}}{1}
      {9}{{{\color{blue}9}}}{1}
      {:}{{{\color{red}:}}}{1}
      {,}{{{\color{red},}}}{1}
      {"}{{{\color{orange}"}}}{1}
      {[}{{{\color{green}[}}}{1}
      {]}{{{\color{green}]}}}{1}
      {\{}{{{\color{green}\{}}}{1}
      {\}}{{{\color{green}\}}}}{1}
}

\usepackage{microtype}
\usepackage{hyperref}
\usepackage{url}
\usepackage{booktabs}
\usepackage{enumitem}
\usepackage{multicol}
\usepackage{multirow}
\usepackage{CJKutf8}
\usepackage{amsmath}
\usepackage{siunitx}
\usepackage{floatflt}
\usepackage{wrapfig}
\usepackage{graphicx}
\usepackage{booktabs}
\usepackage{wrapfig}
\usepackage{authblk}
\usepackage{lipsum}
\usepackage{dsfont}
\usepackage{bbm}
\usepackage{algorithm}
\usepackage{algorithmicx}
\usepackage{algpseudocode}
\usepackage{microtype}
\usepackage{graphicx}
\usepackage{subcaption}
\usepackage{multirow}
\usepackage{booktabs} 
\usepackage{pifont}  
\usepackage{graphicx}  
\usepackage{subcaption} 
\usepackage{hyperref}
\usepackage{pgfplotstable}
\usepackage{makecell}
\usepackage[utf8]{inputenc}
\usepackage[most]{tcolorbox}
\newtcolorbox{alprompt}[1]{
        boxrule = 1pt,
        fontupper = \small\tt,
        fonttitle = \bf\color{black},
        arc = 2pt,
        rounded corners,
        colframe = black,
        colbacktitle = white!97!yellow,
        colback = white!97!yellow,
        title = #1,
}
\definecolor{darkgreen}{rgb}{0.0, 0.5, 0.0}
\definecolor{darkgray}{gray}{0.4}
\definecolor{maroon}{rgb}{0.5, 0.0, 0.0}
\definecolor{navy}{rgb}{0.0, 0.0, 0.5}
\definecolor{teal}{rgb}{0.0, 0.5, 0.5}

\definecolor{deepblue}{RGB}{41, 128, 185}

\definecolor{mylightgreen}{RGB}{144,238,144}
\definecolor{mylightblue}{RGB}{173,216,230}

\definecolor{outerboxcolor}{gray}{0.90} 
\definecolor{innerboxcolor}{rgb}{1,1,1}

\definecolor{nred}{RGB}{196, 38, 11}
\definecolor{ngreen}{RGB}{18, 141, 21}
\definecolor{nblue}{RGB}{41, 52, 190}

\usepackage{listings}
\usepackage{xcolor}
\usepackage{tcolorbox}
\tcbset{listing style/.style={sharp corners,boxrule=0.5pt}}
\tcbuselibrary{skins} 

\algnewcommand{\LeftComment}[1]{\Statex \(\triangleright\) #1}

\usepackage{array}
\usepackage{amsmath}
\usepackage{mathtools}
\usepackage{amsthm}
\usepackage{arydshln}
\usepackage[capitalize,noabbrev]{cleveref}
\usepackage{adjustbox} 
\usepackage{enumitem}
\usepackage{longtable}
\usepackage{pgfplots}
\usepackage{fancyvrb}
\usepackage{tcolorbox}
\newtcblisting{promptbox}[1][]{%
  enhanced,
  breakable,
  colback=gray!10,
  colframe=gray!50,
  arc=1mm,
  boxsep=4pt,
  fontupper=\ttfamily\small,
  title=#1,
  listing only,
  listing engine=listings,
  listing options={
    inputencoding=latin1,    
    breaklines=true,
    breakatwhitespace=false,
    basicstyle=\ttfamily\small,
    columns=fullflexible,
    escapeinside={(*@}{@*)}      
  }%
}

\theoremstyle{plain}

\theoremstyle{definition}

\theoremstyle{remark}

\usepackage[textsize=tiny]{todonotes}

\makeatletter
\renewcommand{\@fnsymbol}[1]{\ensuremath{\ifcase#1\or \dag\or \ddag\or \S\or \P\or \|\or **\fi}}
\makeatother

\colmfinalcopy

\title{Parallel-R1: Towards Parallel Thinking via Reinforcement Learning}

\author[]{Tong Zheng$^{1,2}$\thanks{This work was done during Tong Zheng's internship at Tencent AI Lab Seattle.}, \ Hongming Zhang$^{1}$, Wenhao Yu$^{1}$, Xiaoyang Wang$^{1}$, Runpeng Dai$^{1,3}$, \\ Rui Liu$^{2}$, Huiwen Bao$^{4}$, Chengsong Huang$^{5}$, Heng Huang$^{2}$, Dong Yu}

\affil[ ]{$^{1}$Tencent AI Lab Seattle, $^{2}$University of Maryland, College Park, }
\affil[ ]{$^{3}$University of North Carolina at Chapel Hill,}
\affil[ ]{$^{4}$City University of Hong Kong, $^{5}$Washington University in St. Louis}

\begin{document}

\maketitle

\begin{abstract}

Parallel thinking has emerged as a novel approach for enhancing the reasoning capabilities of large language models (LLMs) by exploring multiple reasoning paths concurrently. However, activating such capabilities through training remains challenging, as existing methods predominantly rely on supervised fine-tuning (SFT) over synthetic data, which encourages teacher-forced imitation rather than exploration and generalization. Different from them, we propose \textbf{Parallel-R1}, the first reinforcement learning (RL) framework that enables parallel thinking behaviors for complex real-world reasoning tasks. Our framework employs a progressive curriculum that explicitly addresses the cold-start problem in training parallel thinking with RL. We first use SFT on prompt-generated trajectories from easier tasks to instill the parallel thinking ability, then transition to RL to explore and generalize this skill on harder problems. Experiments on various math benchmarks, including MATH, AMC23, and AIME, show that Parallel-R1 successfully instills parallel thinking, leading to 8.4\% accuracy improvements over the sequential thinking model trained directly on challenging tasks with RL. Further analysis reveals a clear shift in the model's thinking behavior: at an early stage, it uses parallel thinking as an exploration strategy, while in a later stage, it uses the same capability for multi-perspective verification. Most significantly, we validate parallel thinking as a \textbf{mid-training exploration scaffold}, where this temporary exploratory phase unlocks a higher performance ceiling after RL, yielding a \textbf{42.9\%} improvement over the baseline on AIME25.  Our model, data, and code will be open-source at \url{https://github.com/zhengkid/Parallel-R1}.

\end{abstract}

\section{Introduction}

\begin{figure}[t!]
    \centering
    \includegraphics[width=\linewidth]{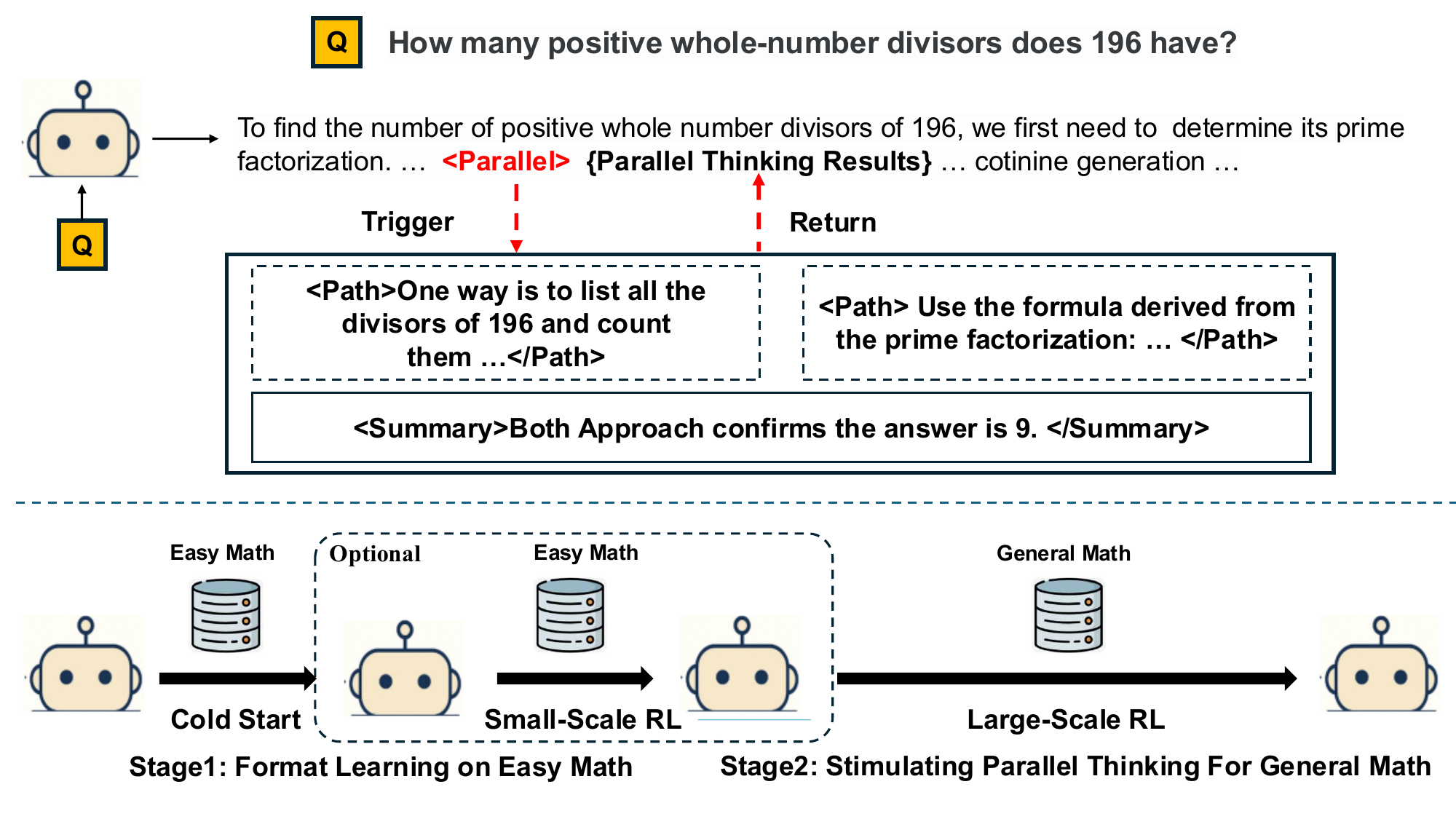}
    \vskip -0.1in 
    \caption{An overview of the proposed framework. (Top) During inference, the model generates in a standard auto-regressive fashion until it emits a special \texttt{<Parallel>} tag. At that point, it spawns multiple threads to explore different solution paths or perspectives, then summarizes their outputs. These contents are merged back into the main context, and generation continues. This cycle may repeat several times before the model arrives at the final answer. (Bottom) Parallel thinking ability is obtained by a progressive multi-stage training approach. Intuitively, the approach first equips the model with parallel thinking ability on easy math problems and then progressively extends it to more general and difficult problems through reinforcement learning.
    }
    \label{fig:overview-framework}
\end{figure}

Google’s Gemini recently credited its success at the International Mathematical Olympiad in part to a new capability: parallel thinking~\citep{deepmind2025gemini}. 
This approach, as exemplified by Figure \ref{fig:overview-framework} (top), involves jointly conducting both parallel and sequential thinking. 
This success highlights the value of parallel thinking as more than a technical trick. Indeed, cognitive science suggests that humans often engage in such thinking, considering multiple possibilities simultaneously before synthesizing them into coherent conclusions. This process encourages divergent thought, prevents premature ``lock-in'' to a single, potentially suboptimal solution, and facilitates structured, deliberate reasoning~\citep{clark1989microcognition, jackendoff2011parallel}. Inspired by these, we investigate how to effectively instill parallel thinking in large language models (LLMs).

Despite its potential, the question of how to activate parallel thinking remains open. While test-time strategies~\citep{yao2023tree, wang2022self, brown2024large, zhang2024accessing, hsu2025group, rodionov2025hogwild, fu2025deep} can elicit such behavior at the cost of high inference overhead, there is a growing interest in permanently instilling this capability through training. However, current training-based approaches fall short of this goal. Methods based on supervised fine-tuning (SFT)~\citep{yang2025multiverse,macfarlane2025instilling,chen2025aspd}, for instance, essentially perform behavioral cloning on pre-generated reasoning trajectories. This approach often relies on complex and costly data pipelines to synthesize high-quality parallel thinking data, leading to superficial pattern matching rather than the acquisition of a deep, intrinsic reasoning skill. Consequently, while models can replicate known patterns, their ability to generalize the underlying parallel thinking strategy is severely limited. 

In contrast, reinforcement learning (RL) offers a more scalable approach to activating the parallel thinking ability of LLMs since we could let the model explore and learn such behaviors in the wild.
However, applying RL to teach models to conduct parallel thinking is not trivial. 
Since the current LLMs have not seen parallel thinking behavior during the pre-training and sft, they cannot generate such trajectories during explorations for the model to learn from.
Thus, the cold-start training becomes crucial. 
The goal of this stage is to teach the model basic formats without harming it too much, which requires a small-scale, high-quality dataset. However, the fact is that high-quality parallel thinking data for complex, real-world problems is extremely rare in natural text and difficult to synthesize. This explains why successful applications of RL for parallel thinking have been confined to narrow, synthetic domains, such as the CountDown task~\citep{pan2025learning}.
Additionally, the best reward function for RL remains unclear. If we only use the final correctness as the reward, the model might take shortcuts to forget the complex but better parallel thinking strategy. On the other hand, if we force the model to use a thinking strategy, the model might learn to use parallel thinking in unnecessary scenarios.
Lastly, the strategic role and underlying mechanisms of parallel thinking in LLMs are largely a black box. Even if a model acquires this ability, critical questions remain unanswered. For instance, how does the model's strategy evolve throughout training? Without understanding this dynamic, it's impossible to fully unlock the potential of parallel thinking technology.

To address these challenges, we present the \textbf{first reinforcement learning framework} designed to help models to learn parallel thinking behavior via exploration on general mathematical reasoning tasks. First, to resolve the critical cold-start problem, we propose a progressive curriculum. As shown in Figure \ref{fig:overview-framework}, it begins with supervised fine-tuning on simpler problems, for which we find high-quality parallel thinking data can be generated easily via simple prompting (see Table \ref{tab:quality-wrap}). This initial stage, using our created \textit{Parallel-GSM8K} dataset, effectively teaches the model the basic format of parallel thinking before it transitions to reinforcement learning on more difficult tasks to explore and generalize this new ability. Second, we tackle the critical challenge of reward design by exploring how to balance final accuracy with the desired parallel thinking structure. We propose and investigate multiple reward schemes. Our key finding is an effective alternating reward strategy, which switches between an outcome-based (accuracy) reward and a reward that encourages parallel thinking behaviors within fixed windows, e.g., every 10 steps. We show this approach achieves a superior balance between high performance and consistent utilization of parallel thinking compared to using a single reward type alone. Lastly, to open the "black box" of its strategic role, we conduct a detailed analysis of the model's learned behavior throughout the training process. Our analysis reveals a clear strategic evolution: the model initially leverages parallel paths for computational exploration to discover potential solutions, but as it gains proficiency, its strategy shifts towards using them for multi-perspective verification to confirm the final answer. This finding provides the first empirical evidence of how an LLM's reasoning strategy with parallel thinking evolves, offering crucial insights into the underlying mechanisms that drive its effectiveness. Based on this, we further conceptualize and empirically validate the idea of using parallel thinking as a mid-training exploration scaffold—a temporary exploratory phase that unlocks a higher performance ceiling, notably achieving a peak accuracy of \textbf{25.6\%} on the challenging AIME25 benchmark. We investigate these contributions across both causal and structured model variants to provide robust insights into architectural design.

In all, our core contributions can be summarized as follows:
\begin{itemize}[itemsep=0.0pt,topsep=0pt,leftmargin=*]
    \item We present the \textbf{first RL framework} to learn parallel thinking from scratch on general mathematical reasoning tasks, enabled by our \textbf{progressive curriculum} and dedicated \textbf{Reward Design}.
    \item We provide a deep analysis of the learning dynamics, revealing that the model's strategy evolves from \textbf{exploration to verification}. We further identify and empirically validate the concept of parallel thinking as a \textbf{mid-training exploration scaffold}.
    \item We provide comprehensive empirical validation, including a comparison of \textbf{causal and structured model variants}. Our approach yields consistent gains across multiple benchmarks, and ablations offer practical insights into reward and architectural design.
\end{itemize}

\section{Related Work}
\subsection{Parallel Thinking}
Parallel thinking has emerged as an active area of research recently~\citep{yao2023tree, wang2022self, brown2024large, zhang2024accessing, huang2025efficient, pan2025learning, huang2024divide,hsu2025group, rodionov2025hogwild, yang2025multiverse,jin2025learning}. 
Among them, a common brute-force strategy is to spawn multiple independent trajectories at the very beginning and join their outcomes only at the end \citep{brown2024large, wang2022self}, or to exchange thoughts at fixed intervals \citep{rodionov2025hogwild, hsu2025group}. Obviously, such schemes lack adaptivity as the points of branching and aggregating are dictated by a pre-defined schedule, not conditioned on the intermediate progress of the thinking process itself. To achieve finer-grained control, methods such as Monte Carlo Tree Search \citep{zhang2024accessing} and Tree of Thoughts \citep{yao2023tree} offer more nuanced parallelism; however, they are still guided by hand-crafted heuristics based on external verifiers. More recent work \citep{pan2025learning, yang2025multiverse} strives for adaptivity through RL or SFT. However, these studies either (i) focus mainly on efficiency—losslessly converting a single long chain-of-thought into an adaptive parallel form via SFT, which limits the discovery of new reasoning patterns, or (ii) demonstrate RL only on toy tasks such as Countdown. In this work, we argue that learning parallel thinking via RL is a more generic and promising direction: it not only retains efficiency but also uncovers novel, highly adaptive reasoning behaviors, leading to improved performance beyond the ``lossless transformation'' paradigm of \citet{yang2025multiverse}. To this end, we proposed the first RL framework to stimulate adaptive parallel thinking for general mathematical tasks.

\subsection{Improving Reasoning via RLVR}
Reinforcement Learning with Verifiable Rewards (RLVR) optimizes language models via reinforcement learning using outcome-based, automatically checkable rewards, eliminating the need for trained reward models or step-level human annotations. Recent advances have demonstrated RLVR’s effectiveness across diverse domains—including mathematical problem solving~\citep{guo2025deepseek}, coding ~\citep{wang2025code}, multi-modal reasoning~\citep{huang2025vision, wang2025vl,zheng2025learning,li2025self}, relation extraction~\citep{dai2025r1}, and interactive GUI navigation~\citep{shi2025mobilegui}. In parallel, a growing body of work aims to make RLVR more efficient and stable, proposing new training paradigms such as self-play~\citep{liu2025spiral, huang2025r} and test-time RL~\citep{zuo2025ttrl}, as well as more robust RL algorithms including DAPO~\citep{yu2025dapo}, VAPO~\citep{yue2025vapo}, and high entropy guided optimization~\citep{wang2025beyond}. However, important challenges remain. Existing methods often leave unresolved issues of faithfulness~\citep{tanneru2024hardness, chen2025reasoning, zhou2025dissecting} and robustness \citep{sabbaghi2025adversarial, dai2025breach}. Moreover, most approaches adopt a strictly sequential reasoning paradigm.  This limitation is fundamental, as LLMs do not inherently possess parallel thinking capabilities, making it a significant and unresolved challenge to instill this skill using standard RLVR methods. To address this specific challenge, we introduces the first reinforcement learning framework that leverages a progressive curriculum to effectively instill the parallel thinking ability in LLMs, fundamentally enhancing their intrinsic reasoning capabilities.

\section{Learning Parallel Thinking via Reinforcement Learning}
\subsection{Overview}

Previous methods for training parallel thinking, such as those in~\citep{yang2025multiverse,macfarlane2025instilling,chen2025aspd}, primarily rely on SFT, a paradigm that suffers from several key limitations. By its nature, SFT's success is entirely dependent on the quality of pre-generated training data. This creates a critical dependency on complex and costly data pipelines, especially when generating data for final, challenging problems. Furthermore, this approach constrains the model to merely mimicking known patterns, which hinders the acquisition of a deep, generalizable reasoning skill.
To overcome these limitations, we introduce a reinforcement learning (RL) framework.

The key insight of our approach is to bypass the need for the complex data pipelines often considered essential for generating training data on final challenging problems~\citep{yang2025multiverse,macfarlane2025instilling}. Instead, we generate high-quality 'cold-start' data by using simple tasks, and then leverage this data to enable the model to learn parallel thinking on much harder problems via reinforcement learning.
We then explore two distinct settings for learning parallel thinking via RL: without architectural modifications and with architectural modifications. Specifically, the latter involves modifying the model's self-attention mask and position ids to prevent cross-attention between parallel reasoning paths, thereby enforcing their structural independence. 

In the subsequent sections, we first define our parallel thinking behaviors and their inference workflow. We then describe our data pipeline for generating high-quality training data. Finally, we present our RL training recipes for both settings.

\subsection{Formulation of Parallel Thinking Behaviors}\label{sec:format}
Intuitively, in human problem-solving, we often encounter moments of confusion or uncertainty, which are referred to as "critical steps" within a reasoning chain. At these points, engaging in parallel thinking allows us to explore multiple solution paths simultaneously and converge on a higher-quality conclusion. Inspired by human problem-solving patterns, we formalize LLM's parallel thinking in two stages:
\begin{itemize}
    \item[1.] \textbf{Exploration}: When the model detects a critical step, it temporarily suspends the main chain and launches a multi‑thread search, generating \(N\) independent trajectories simultaneously.
    \item[2.] \textbf{Summary}: After exploration, the model aggregates the outcomes, distills key insights, and resolves conflicts to arrive at the most promising conclusion. It then automatically resumes the main reasoning chain with the summarized conclusion.
\end{itemize}
We allow the model to repeat these two phases whenever needed during the reasoning process. We illustrate this process in Figure \ref{fig:overview-framework} (Top). To implement this behavior, we introduce three control tags, \verb|<Parallel>…</Parallel>|, \verb|<Path>…</Path>|, and \verb|<Summary>…</Summary>|, which correspond to the exploration phase, the isolation of reasoning threads, and the summary of the parallel thinking, respectively. With these tags, we can define the workflow at the inference phase as follows:

\paragraph{Workflow at Inference Phase} At inference time, our model dynamically executes the parallel thinking behaviors as follows: It first conducts auto-regressive generation in the main reasoning process. Whenever it predicts a \verb|<Parallel>| token, it pauses the main reasoning chain and concurrently expands multiple reasoning threads within separate \verb|<Path>...</Path>| blocks. After generating all parallel threads, the model automatically aggregates their outputs into a concise \verb|<Summary>...</Summary>| block, integrating insights from diverse perspectives. Finally, all contexts of parallel thinking are used to resume and complete the main reasoning path. Such adaptive and dynamic parallel inference effectively leverages parallelism.

\subsection{The Simple and Scalable Data Pipeline for Parallel Thinking}
\label{sec:data_pipeline}
\begin{tcolorbox}[title=\textbf{Key Finding 1}]
A powerful model can produce valid parallel-thinking traces for 83.6\% of simple GSM8K problems, but fails to generate a single valid trace for the more challenging DAPO problems (0.0\% success rate).
\end{tcolorbox}

Collecting high-quality parallel thinking data is a significant challenge. Even though humans think in the parallel fashion, they will summarize and only say/write the summarization. Thus, such data is extremely rare in the natural distribution. Existing approaches, such as the one described in \citep{yang2025multiverse}, try to solve this by leveraging the inherent parallelism of long CoT reasoning chains. However, these methods rely on complex, multi-stage data pipelines that, while avoiding costly human annotations, are computationally intensive and fundamentally limited in their scalability.

Our approach is based on a key finding from our preliminary experiments. We found that while a simple prompting approach struggles to generate high-quality parallel-thinking data for complex problems like DAPO, it proves highly effective for simpler tasks like GSM8K. The data in Table \ref{tab:quality-wrap} supports this finding. Based on this discovery, we propose a simple and scalable data pipeline that uses detailed zero-shot prompts to construct a large, high-quality corpus for these easier problems. 

As the structured model variant (described in Section \ref{sec:structure}) utilizes architectural modifications like path-window attention masks, it requires strict format adherence for successful training. Therefore, to ensure the quality and alignment of this corpus, we perform an additional filtering step, a Parallel Thinking Format Check, which is implemented by Algorithm \ref{alg:format-check}.  Crucially, we make the strategic choice to use this 'cold-start' data not to teach the model how to solve the final target tasks, but specifically to teach it the format of parallel thinking. This initial stage allows us to transition from a data-intensive approach to a more efficient reinforcement learning framework that can learn to elicit and strengthen parallel thinking behaviors from the ground up.

\begin{table}[t]
\centering
\caption{Comparison of Parallel-Thinking Data Quality Generated by DeepSeek-R1-0528-Qwen-3-8B on DAPO and GSM8K under identical prompts and sampling settings. The results show that, with simple prompting, state-of-the-art models still struggle to produce concise, high-quality parallel reasoning traces for challenging mathematics problems.}
\vspace{-0.1in}
\label{tab:quality-wrap}
\begin{tabular}{@{}lcc@{}}
\toprule
\textbf{Data} & \textbf{\# Samples} & \textbf{Parallel Thinking Format (\%)}  \\ \midrule
GSM8K & 7472 & \bf 83.7 \\
DAPO & 17916 & 0.0    \\
\bottomrule
\end{tabular}
\end{table}

\subsection{Eliciting Parallel Thinking via Reinforcement Learning in Causal Models}
Unlike prior approaches that use complex and costly data pipelines, we design a simple and scalable data pipeline to efficiently generate a large, high-quality parallel thinking dataset on easy math problems. This dataset serves as a crucial cold start to teach the model the correct format for parallel thinking. Our key idea is to use an RL framework to generalize this format and ability from simple problems to more difficult mathematical tasks. In this section, we explore strategies to elicit this parallel thinking behavior without modifying the model's architecture.

\subsubsection{Reinforcement Learning Algorithms}\label{sec:grpo}

We use Group Relative Policy Optimization (GRPO)~\citep{shao2024deepseekmath} as our reinforcement learning algorithm. Let \(q\) be a question, and let \(\{o_i\}_{i=1}^G\) be \(G\) candidate responses sampled from the old policy \(\pi_{\theta_{\mathrm{old}}}(\cdot\!\mid q)\). We denote \(r_i\) as the reward for \(o_i\). We define:
\[
\rho_i = \frac{\pi_{\theta}(o_i\!\mid q)}{\pi_{\theta_{\mathrm{old}}}(o_i\!\mid q)}, 
\qquad
\overline{r} = \frac{1}{G}\sum_{j=1}^G r_j,
\qquad
A_i = \frac{r_i - \overline{r}}
           {\sqrt{\tfrac{1}{G}\sum_{j=1}^G (r_j - \overline{r})^2} + \varepsilon_{\mathrm{stab}}},
\]
where \(\varepsilon_{\mathrm{stab}}\) is a constant for numerical stability and  $A_i$ is the advantage. The GRPO loss is then:
\[
\mathcal{L}_{\mathrm{GRPO}}(\theta)
= 
\mathbb{E}_{\substack{q\sim\mathcal{D}\\\{o_i\}\sim\pi_{\theta_{\mathrm{old}}}}}
\Biggl[
\frac{1}{G}\sum_{i=1}^G
\min\!\bigl(\rho_i A_i,\;\mathrm{clip}(\rho_i,1-\alpha,1+\alpha)\,A_i\bigr)
\;-\;\beta\,D_{\mathrm{KL}}\bigl(\pi_{\theta}\,\Vert\,\pi_{\mathrm{ref}}\bigr)
\Biggr].
\]

Our model's rollout process follows a multi-turn interactive framework where the LLM alternates between autoregressive generation, parallel exploration, and summarization. The model generates a reasoning prefix until it predicts a \verb|<Parallel>| block, within which it spawns several independent <Path> segments. Once all paths are completed, it produces a \verb|<Summary>| segment that integrates insights from the parallel paths into a coherent continuation. This process repeats as needed.

\subsubsection{The Training Recipe and Reward Modeling}
The overall training recipe consists of three stages: 1) Cold-Start Stage; 2) RL on Easy Math, and 3) RL on General Math.

\paragraph{Cold-Start Stage} We construct and collect a small set of parallel-thinking format examples to fine-tune the initial RL actor using the approach in Section \ref{sec:data_pipeline}. Specifically, we use a distilled Qwen3‑8B model (i.e., DeepSeek-R1-0528-Qwen-3-8B) to produce high‑quality parallel‑thinking outputs, extracting only non-thinking parts (final short CoT) as gold annotations. 

We select the GSM8K training set, which consists of approximately 7k samples, as the seed dataset. We call the resulting cold-start dataset \textit{Parallel-GSM8K}. This cold-start training is used to teach model the basic format of parallel thinking. 

\paragraph{RL on Easy Math} After the cold start with SFT, the model already possesses the basic ability to generate the tags for parallel thinking, but the behavior is not stable since this special token has never appeared in the pre-training.
To address this issue, we further perform small-scale reinforcement learning to enhance the format learning. In this stage, we use the same question set as the cold-start data 
and use Group Relative Policy Optimization (GRPO)~\citep{shao2024deepseekmath} for our RL training. To ensure parallel ratio and accuracy, the final reward format in this stage is: $R_{final}=R_{\langle\mathrm{Parallel}\rangle} \times R_{\mathrm{acc}}$. Here, the Accuracy Reward ($R_{\mathrm{acc}}$) evaluates the correctness of the final response, while the Parallel Reward ($R_{\langle\mathrm{Parallel}\rangle}$) incentivizes the model to use parallel reasoning paths. This reward structure is designed to be binary and strict: a positive reward of +1 is given only if the generated output contains at least one parallel thinking unit \textbf{AND} the final answer is correct. Otherwise, the model receives a penalty of -1. 

\paragraph{RL on General Math} After the initial training, the model can stably generate control tags and produce outputs in the correct parallel thinking format if needed, but it still struggles with more challenging mathematical tasks. To address this, we apply reinforcement learning to general math datasets, thereby generalizing the model's parallel thinking ability beyond simple cases.

Specifically, we use the same GRPO algorithm introduced in Section \ref{sec:grpo} with accuracy reward ($R_{\mathrm{acc}}$) as our sole reward. This is because the primary goal of this stage is to improve task performance. For the seed problems, we choose the widely used DAPO dataset \citep{yu2025dapo}. Finally, the models produced by this stage are our \textit{Parallel-Seen} variants.

\subsection{Eliciting Parallel Thinking via Reinforcement Learning in Structure Models}
\label{sec:structure}
In the previous section, we explored an RL framework that trains models to use parallel thinking without modifying their underlying architecture. However, this approach, which we call Parallel-Seen, does not explicitly isolate reasoning paths. As a result, hidden representations from one path can inadvertently leak into others, and gradients across paths can interfere with each other during training.

\begin{wrapfigure}[16]{r}{0.43\linewidth}
\vspace{-2em}
    \centering
    \includegraphics[width=\linewidth]{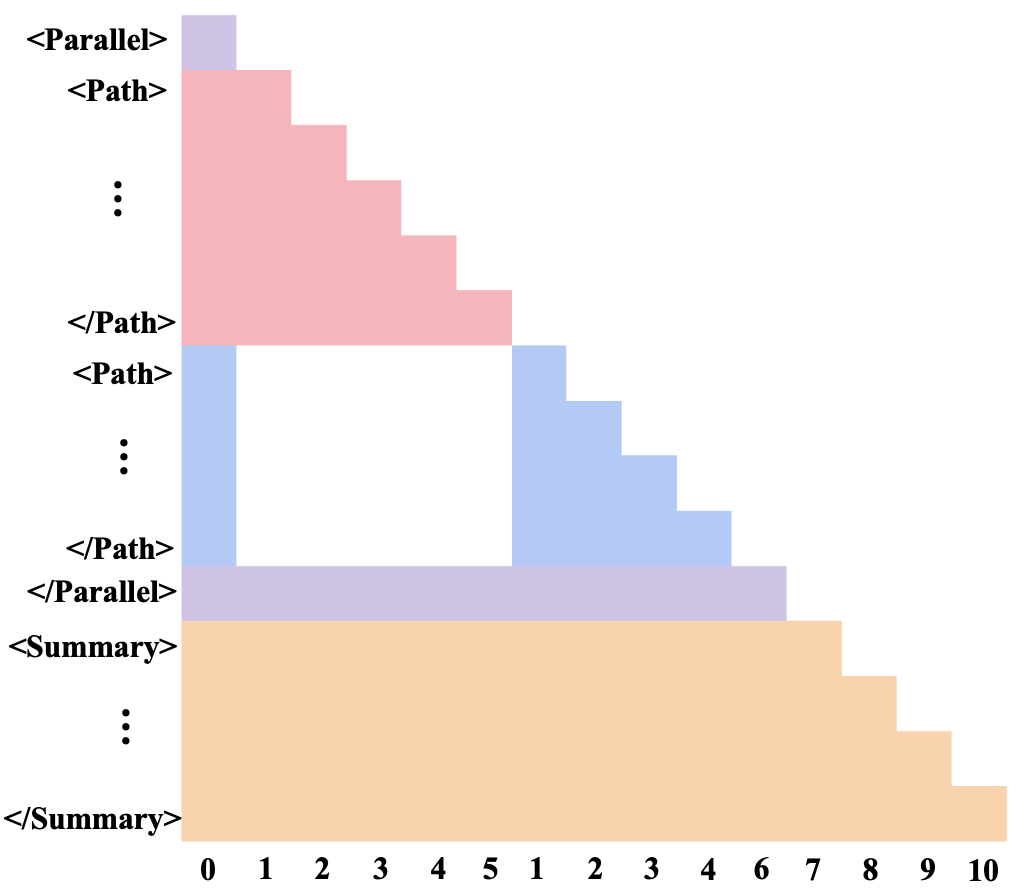}
    \vspace{-0.2in}
    \caption{Illustration of the structured attention mask and position IDs, where different paths and the summary block have distinct visibility regions. Blank regions indicate tokens that cannot attend to each other, while colored regions indicate tokens that can.}
    \label{fig:attention_mask}
\end{wrapfigure}

To explore an alternative solution, we introduce a structured variant of our framework, Parallel-Unseen. This model incorporates explicit inductive biases into the attention mechanism to enforce path isolation. Specifically, inspired by prior work \citep{yang2025multiverse}, we design path-window masking and multiverse position encodings to achieve the goal.

\subsubsection{Structured Attention Mechanism}
We incorporate these inductive biases directly into the attention layer, as shown in Figure \ref{fig:attention_mask}.
\begin{itemize}
\item Path-window masking restricts each token within a \verb|<Path>| block to attend only to tokens from the same path and the shared context. This prevents cross-path information leakage.

\item Multiverse position encodings assign a disjoint set of position indices to each path, ensuring that the positional embedding space does not overlap. 
\end{itemize}
Together, these constraints enforce explicit isolation among reasoning threads while preserving visibility from the shared \verb|<Summary>| block, which is essential for integrating insights across paths.

\subsubsection{The Training Recipe and Reward Modeling}
In preliminary experiments, we find that directly applying the progressive training recipe from Parallel-R1-Seen to the structured variant proves ineffective. We attribute this to the poor generalization of attention masks from easy to hard math~\citep{yang2025ape}. To address this limitation, we remove the stage one RL and redesign the reward schedule and evaluate two alternative schemes.
\paragraph{(S1) Accuracy-only.} 
We optimize solely for task correctness, removing direct incentives for parallel usage. 

\paragraph{(S2) Alternating accuracy and parallel.} 
In this scheme, we alternate between two different rewards within fixed windows of W=10 steps. For 80\% of the steps, we use a standard accuracy-only reward ($R_{\mathrm{acc}}$). For the remaining 20\% of the steps, we use a tiered reward system to provide a nuanced incentive for parallel thinking:
\begin{itemize}
    \item +1.2: If the generated output contains at least one parallel thinking unit AND the final answer is correct.

\item  +1.0: If the generated output does not contain a parallel thinking unit AND the final answer is correct.

\item  -1.0: For all other cases, including incorrect answers.
\end{itemize}
This schedule reintroduces a calibrated incentive for parallel usage without letting it dominate training. Together, these reward designs allow Parallel-R1-Unseen to leverage the benefits of structural isolation while avoiding overfitting to superficial parallel patterns.

\section{Experiments}
\subsection{Experimental Setups}
\paragraph{Model.} We use Qwen‑3‑4B‑Base~\citep{yang2025qwen3} as our backbone, the latest state‑of‑the‑art open‑source model at this scale, offering an ideal balance between performance and efficiency.
\paragraph{Evaluation.} We measure our models on four standard mathematical reasoning benchmarks, including AIME'24, AIME'25, AMC'23, and MATH~\citep{hendrycks2021measuring}. On the MATH dataset, we generate one response per question using a sampling temperature of 
$T=1.0$. For the remaining three datasets, we sample 16 independent responses per question at the same temperature and report the average accuracy (i.e., $\mathrm{mean}@16$) to reduce randomness, which is consistent with settings in \citet{wang2025beyond}. We additionally report $\mathrm{pass}@16$ to show the upper bound of our approach.

\paragraph{Training Details.} Our codebase is adapted from VERL~\citep{sheng2024hybridflow}, where we primarily follow its official training recipe without any hyperparameter tuning. In the cold start stage, we perform SFT on our curated Parallel-GSM8K, using a batch size of 128, a learning rate of 1e-5, a weight decay of 0.01, and a warm-up step ratio of 0.1 with the cosine learning‐rate schedule, resulting in 58/230 gradient update steps for Parallel-SFT-Seen and Parallel-SFT-Uneen, respectively. For Stage 1, we optionally perform RL on GSM8K for five epochs, using a batch size of 1024, 5 rollouts, and a learning rate of 1e-6 without warm-up or learning rate scheduling, resulting in 35 gradient update steps. For Stage 2, we perform RL on the DAPO training set for 300 gradient update steps, using a batch size of 512, a rollout of 8, and a learning rate of \mbox{1e-6 without warm-up or learning rate scheduling.}

\begin{table}[t!]
\centering
\caption{Performance comparison on mathematical reasoning benchmarks for the Qwen-3-4B-Base model trained under different parallel thinking configurations. 
We report Mean@16 and Pass@16 for AIME25, AIME24, and AMC23, while MATH is evaluated with Mean@1.}
\vspace{-0.1in}
\small
\setlength{\tabcolsep}{2pt} 
\resizebox{\textwidth}{!}{\begin{tabular}{lccccccccc}
\toprule
\multirow{2}{*}{\textbf{Method}} & \multirow{2}{*}{\textbf{\# Parallel}} 
& \multicolumn{2}{c}{\textbf{AIME25}} 
& \multicolumn{2}{c}{\textbf{AIME24}} 
& \multicolumn{2}{c}{\textbf{AMC23}} 
& \multirow{2}{*}{\textbf{MATH}} 
& \multirow{2}{*}{\textbf{Avg.}} \\
\cmidrule(lr){3-4}\cmidrule(lr){5-6}\cmidrule(lr){7-8}
& & Mean@16  & Pass@16 
  & Mean@16  & Pass@16 
  & Mean@16  & Pass@16 
  & Mean@1 & \\
\midrule
Qwen3-4B-Base & 0.0 & 1.3  & 10.2 & 2.9 & 16.5 & 8.1  & 51.2 & 13.9 & 6.6 \\
\midrule
\multicolumn{10}{l}{\textit{SFT + Parallel}} \\
Parallel-SFT-Seen & 95.6 & 8.0 & 29.8 & 10.6 & 26.4 & 48.9 & 79.2 & 76.6 & 36.0\\
Parallel-SFT-Unseen & 95.6 & 5.2 & 20.9 & 8.5 & 26.7 & 41.7 & 80.1 & 71.5 &  31.7 \\
\midrule
\multicolumn{10}{l}{\textit{RL Approach}} \\
GRPO (DAPO) & 0.0 & 14.8  & 32.4 & 18.5  & 30.6 & 63.6  & 85.1 & 83.5 & 45.1 \\
\quad + RL on GSM8K & 0.0 & 13.3  & 26.3 & 18.8  & 34.9 & 66.4  & 82.2 & 82.6 & 45.3 \\
Parallel-R1-Seen & 27.3 & \bf 19.2  & 38.9 & \bf 19.4 & \bf 37.1 & \bf 70.5  & 85.0 & \bf 86.7 & \bf 48.9 \\
Parallel-R1-Unseen (S1) & 13.6 & 17.7  & 37.8 & 18.3  & 33.2 & 69.7 & 88.9 & 82.6 & 47.1 \\
Parallel-R1-Unseen (S2) & \bf 63.0 & 19.0 & \bf 42.2 & 16.3  & 31.8 & 67.5  & \bf 91.5 & 84.5 & 46.8 \\
\bottomrule
\end{tabular}}
\label{tab:adaptive_parallel}
\end{table}

\subsection{Main Results}

Table \ref{tab:adaptive_parallel} presents the results across four benchmarks: AIME25, AIME24, AMC23, and MATH. We compare our method against two baselines: 1) RL with GRPO algorithm directly on the DAPO training set, and 2) RL with GRPO in two stages: first trained on the GSM8K data, then further trained with RL on the DAPO training set. The second baseline is included to ensure fair comparison.

Our progressive \texttt{Parallel-R1} framework proved to be the most effective approach, consistently outperforming all baselines as shown in Table~\ref{tab:adaptive_parallel}. The top-performing causal variant, \texttt{Parallel-R1-Seen}, achieved the highest average score of 48.9. This success stems from a curriculum designed to overcome the limitations of simpler methods. For instance, while SFT provides a substantial foundational improvement (e.g., 31.7 for \texttt{Parallel-SFT-Unseen} vs. 4.6 for the base model), it is insufficient for advanced reasoning and falls considerably short of the standard GRPO baseline's score of 45.1. Besides, we found that a naive additional RL on easier data offers only a marginal benefit on average (45.3 vs. 45.1), validating our strategy of using cold start for targeted format and behavior learning.

Our results also reveal key design trade-offs. The superior performance of the Seen model compared to its structured counterparts suggests that explicit architectural modifications can be detrimental to RL training. Furthermore, the comparison between reward schedules for \texttt{Parallel-R1-Unseen (S1)} and \texttt{(S2)} highlights that reward design is essential for effectively managing the trade-off between the parallel ratio and overall performance. We provide detailed analysis in Section \ref{sec:Ablation_Studies_Reward}.

\subsection{Analysis}

\begin{table*}[t!]
\caption{Ablation Study on Training Approach: Comparison of different training configurations.}
\vspace{-0.1in}
\label{tab:training-approach-ablation}
\centering
\setlength{\tabcolsep}{1mm}{\begin{tabular}{lccccc}
\toprule
\textbf{Training Configuration} & \textbf{AIME25} & \textbf{AIME24} & \textbf{AMC23} & \textbf{MATH} & \textbf{Avg.} \\
\midrule
\multicolumn{4}{l}{\textbf{Effect of Training Stages}} \\
Parallel-R1-Seen                          &  \bf 19.2 & \bf 19.4 & \bf 70.5 & \bf 86.7 & \bf 48.9 \\
\quad - w/o RL on GSM8K                   &  17.9 & 19.0 &  65.0  & 84.5 & 46.6 \\
Parallel-R1-Unseen (S1)                          &   17.7 & 18.3 & 69.7 & 82.6 & 47.1 \\
\quad + with RL on GSM8K                   &  14.4 & 12.9 &  52.3 & 74.4 & 38.5 \\
\midrule
\multicolumn{4}{l}{\textbf{Effect of Parallel Thinking Prompt}} \\
Parallel-R1-Seen                          &    19.2 & \bf 19.4 & \bf 70.5 & \bf 86.7 & \bf 48.9 \\
\quad - w/o Parallel Thinking Prompt              & \bf 20.4 & 16.5  & 66.7 & 84.8 & 47.1  \\
\bottomrule
\end{tabular}}
\end{table*}

\subsubsection{Ablation on Training Approach}
We further investigate the role of two-stage RL in our training pipeline. One natural question is whether learning on GSM8K, which is a relatively simple math dataset, truly benefits from the RL, given that the structural parallel reasoning format (e.g., the correct use of \texttt{<Parallel>}, \texttt{<Path>}, and \texttt{<Summary>} tokens) can be directly acquired through SFT~\citep{yang2025multiverse}.

Table~\ref{tab:training-approach-ablation} presents the ablation results. For the Causal variant, keeping the Cold Start SFT but removing stage one RL training on GSM8K leads to a consistent performance drop (–2.3\% on average). This indicates that learning format through SFT alone is insufficient. Without stage one RL, the model enters stage two training on general math without having acquired the ability to trigger or use parallel thinking adaptively. As a result, RL training must simultaneously learn both adaptive parallel thinking behavior and mathematical reasoning ability, which is harder to optimize.

Interestingly, the Structure variant exhibits the opposite trend: adding stage one RL on GSM8K severely hurts performance (–8.6\% on average).
We hypothesize that this is because the structured attention mask learned on easy math tasks (GSM8K) does not transfer well to the distribution shift of harder math problems, causing overfitting to superficial patterns, which is consistent with findings in \citep{yang2025ape}.
This contrast highlights a key insight: while stage one RL is crucial for the causal variant to bootstrap adaptive parallel thinking, structural variants require a different training recipe and reward schedule to generalize effectively.


\subsubsection{Ablation on Parallel Thinking Prompt}
We also conduct an ablation study on the effect of our parallel thinking prompt. As shown in Table~\ref{tab:training-approach-ablation}, removing the prompt leads to a performance degradation of up to 1.8\% on average. It indicates that providing more detailed instructions during training helps the model better understand the reasoning process, rather than merely memorizing the output patterns.

\begin{table*}[t!]
\caption{Ablation Study on Reward Modeling for the \textsc{Parallel-R1-Unseen} Model.}
\vspace{-0.1in}
\small
\label{tab:reward-approach-ablation}
\centering
{\setlength{\tabcolsep}{1mm}{\begin{tabular}{lcccccc}
\toprule
\textbf{Training Configuration} & \textbf{Parallel Ratio} & \textbf{AIME 25} & \textbf{AIME 24} & \textbf{AMC 23} & \textbf{MATH}  \\
\midrule

Accuracy  & 13.6 & 17.7 & \bf 18.3 & \bf 69.7 & 82.6 \\
Parallel & \bf 80.3 & 17.7 & 15.2 & 59.4 & 81.7  \\
Alternating Acc./Parallel &  63.0 & \bf 19.0 & 16.3 & 67.5 & \bf 84.5  \\
\bottomrule
\end{tabular}}}
\end{table*}

\subsubsection{Ablation Studies on Reward Modeling: How to Effectively Stimulate Parallel Thinking }
\label{sec:Ablation_Studies_Reward}
In our work, a key question is ``how to effectively stimulate parallel thinking behavior.'' To answer this, we test several reward modeling strategies, including direct accuracy, direct parallel, and an alternating approach. We present the results in Table \ref{tab:reward-approach-ablation}. We can draw the following observations:
\begin{itemize}[itemsep=0.0pt,topsep=0pt,leftmargin=*]
    \item \textbf{Directly rewarding accuracy is insufficient to stimulate parallel thinking}. The "Accuracy" configuration, which optimizes solely for problem correctness, yields the highest performance on two out of four benchmarks, particularly on the AMC dataset (69.7). However, this approach yields a very low parallel ratio of 13.6, indicating that the model, although effective at problem-solving, does not naturally adopt a parallel reasoning style. This suggests that simply pursuing accuracy does not encourage the model to explore alternative reasoning structures.
    \item \textbf{Directly rewarding parallel thinking is detrimental to overall performance}. In contrast, the "Parallel" configuration, which directly rewards the generation of parallel structures, achieves an impressively high parallel ratio of 80.3. This demonstrates that the model is highly responsive to this reward signal. However, this focused optimization leads to a significant performance drop across most benchmarks, with scores on AMC 23 and MATH decreasing to 59.4 and 81.7, respectively. This highlights a critical finding: unconstrained encouragement of parallel behavior can cause the model to prioritize structural form over logical correctness, ultimately hindering its problem-solving capabilities.
    \item \textbf{Our alternating approach effectively balances parallel thinking with performance.} The "Alternating Acc./Parallel" strategy, which periodically switches between rewarding accuracy and parallel structures, provides a superior balance. This approach successfully stimulates parallel thinking, achieving a parallel ratio of 63.0, which is significantly higher than the direct accuracy method. Crucially, this increase in parallel thinking does not come at the expense of performance. In fact, on some challenging benchmarks, such as AIME 25, our method even surpasses the direct accuracy approach, achieving scores of 19.0.

\end{itemize}

\subsection{Evolution of Parallel Thinking Behavior During RL Training}

\begin{tcolorbox}[title=\textbf{Key Finding 2}]
The model's parallel thinking behavior evolves throughout RL training, shifting from early-stage \textbf{computational exploration} to late-stage \textbf{multi-perspective verification}.
\end{tcolorbox}
\begin{wrapfigure}{r}{0.5\textwidth}
  \vspace{-12pt}
  \centering
  \begin{tikzpicture}
    \begin{axis}[
      width=\linewidth, 
      height=6.2cm,
      xlabel={Training Steps},
      ylabel={Relative Position},
      grid=both,
      ymin=0.3, 
      ymax=0.8,
      tick label style={font=\small}, 
      label style={font=\small},
      thick
    ]
      \addplot+[mark=none, red, thick]
        table [x=step, y=mean_position, col sep=comma]
        {Figure/raw_data/parallel_position_trend.csv};
    \end{axis}
  \end{tikzpicture}
  \vspace{-15pt}
  \caption{Dynamics of the relative position of the \texttt{<Parallel>} block during RL training. The increasing trend indicates the model learns to apply parallel thinking later in the reasoning process.}
  \vspace{-10pt} 
  \label{fig:trainin_dynamics_postion_parallel}
\end{wrapfigure}
To better understand how the model's strategy evolves, we analyzed the positional dynamics of the \texttt{<Parallel>} block throughout the RL training. We measured the relative position of each block by dividing its starting token index by the total sequence length of the solution. The training dynamics in Figure~\ref{fig:trainin_dynamics_postion_parallel} show a clear and consistent trend: the average relative position of the \texttt{<Parallel>} block steadily increases as RL training progresses, indicating a strategic shift from applying this feature early in the reasoning chain toward the very end.

We interpret this positional shift as the model adopting a more conservative strategy to maximize its reward, a behavior shaped directly by the final-answer-dominated reward design. In the early stages of training, when the model's reasoning ability is weak, using parallel paths for \textbf{computational exploration} is a necessary, high-variance strategy to discover a potential solution. However, as the model's core reasoning ability improves, such early-stage exploration becomes a liability that could introduce errors and jeopardize the final reward.

Consequently, the model learns a more risk-averse strategy to secure a correct answer. It first derives a solution using a single, high-confidence reasoning path. Only after a potential answer is found, it deploys the \texttt{<Parallel>} block for \textbf{multi-perspective verification}. This late-stage use of parallel thinking confirms the result without risking the integrity of the primary solution path, thus maximizing the probability of receiving a positive reward. This learned behavior aligns with our broader finding of a tension between final-answer optimization and the preservation of diverse reasoning structures.

To further illustrate this behavioral evolution, we present two representative case studies below~(Figure \ref{fig:case_1} and \ref{fig:case_2}). The first case, from an early-stage model, demonstrates the use of parallel thinking for exploration. The second, from the late-stage model, exemplifies the learned, verification-oriented strategy.

\subsection{Extra Bonus: Parallel Thinking as a Mid-Training Exploration Strategy for RL Training}

\begin{tcolorbox}[title=\textbf{Key Finding 3}]
Parallel thinking itself can serve as an effective structured exploration mechanism to improve RL training.
\end{tcolorbox}

In this section, we investigate the hypothesis that \textbf{parallel thinking itself can serve as an effective structured exploration mechanism to improve RL training.} A fundamental challenge in RL is ensuring the model sufficiently explores the policy space to avoid converging to local optima. We posit that by compelling the model to generate multiple, parallel thought blocks at specific reasoning steps, we introduce a strong inductive bias that forces a more structured and diverse exploration, guiding the model toward more robust policy spaces.

\definecolor{ParLine}{HTML}{4C9A6A} 
\usetikzlibrary{shapes.geometric}
\pgfplotstableread[col sep=comma]{Figure/raw_data/stage1.csv}\tblStageOne
\pgfplotstableread[col sep=comma]{Figure/raw_data/stage2.csv}\tblStageTwo

\pgfplotstablesort[sort key=AIME25]\tblStageOneSorted{\tblStageOne} 
\pgfplotstablegetrowsof{\tblStageOneSorted}
\pgfmathtruncatemacro{\StageOneLast}{\pgfplotsretval-1}             
\pgfplotstablegetelem{\StageOneLast}{_step}\of{\tblStageOneSorted}\edef\StageOneMaxX{\pgfplotsretval}
\pgfplotstablegetelem{\StageOneLast}{AIME25}\of{\tblStageOneSorted}\edef\StageOneMaxY{\pgfplotsretval}

\pgfplotstablesort[sort key=AIME25]\tblStageTwoSorted{\tblStageTwo}
\pgfplotstablegetrowsof{\tblStageTwoSorted}
\pgfmathtruncatemacro{\StageTwoLast}{\pgfplotsretval-1}
\pgfplotstablegetelem{\StageTwoLast}{_step}\of{\tblStageTwoSorted}\edef\StageTwoMaxXRaw{\pgfplotsretval}
\pgfplotstablegetelem{\StageTwoLast}{AIME25}\of{\tblStageTwoSorted}\edef\StageTwoMaxY{\pgfplotsretval}
\pgfmathsetmacro{\StageTwoMaxX}{\StageTwoMaxXRaw + 200}              

\begin{figure}[t]
  \centering
  \begin{tikzpicture}

    \begin{axis}[
        name=acc,
        width=0.95\linewidth, height=8cm,
        xmin=0, xmax=550,
        ymin=0, ymax=28,
        xtick distance=100,
        ytick distance=5,
        xlabel={\textbf{Training Steps}},
        ylabel={\textbf{AIME25 (mean@16) (\%)}},
        grid=major,
        grid style={dotted, gray!40},
        tick label style={font=\small},
        label style={font=\bfseries},
        axis y line*=left,   
        ytick pos=left,      
        ylabel style={
    at={(rel axis cs:0,0.5)},  
    yshift=-1em,                
  },
        legend style={fill=white, draw=black!15, rounded corners=2pt,
                      font=\small, at={(0.03,0.97)}, anchor=north west,
                      /tikz/every even column/.style={column sep=0.4em}},
        legend cell align=left,
        every axis plot/.append style={line join=round, line cap=round},
        unbounded coords=discard
    ]

      \addplot+[color=gray!65!black, ultra thick, mark=none]
        table[x=_step, y=AIME25, col sep=comma] {Figure/raw_data/baseline.csv};
      \addplot+[only marks, mark=*, mark size=1.8pt,
                mark options={fill=gray!65!black, draw=white, line width=0.45pt},
                mark repeat=15]
        table[x=_step, y=AIME25, col sep=comma] {Figure/raw_data/baseline.csv};

      \addplot+[color=blue!70!black, ultra thick, mark=none,
                restrict x to domain=0:200]
        table[x=_step, y=AIME25, col sep=comma] {Figure/raw_data/stage1.csv};
      \addplot+[only marks, mark=o, mark size=2pt,
                mark options={fill=white, draw=blue!70!black, line width=0.6pt},
                mark repeat=15, restrict x to domain=0:200]
        table[x=_step, y=AIME25, col sep=comma] {Figure/raw_data/stage1.csv};

      \addplot+[color=red!70!black, ultra thick, mark=none,
                restrict x to domain=200:550]
        table[x expr=\thisrow{_step}+200, y=AIME25, col sep=comma] {Figure/raw_data/stage2.csv};
      \addplot+[only marks, mark=triangle*, mark size=2.2pt,
                mark options={fill=red!60, draw=white, line width=0.6pt},
                mark repeat=15, restrict x to domain=200:550]
        table[x expr=\thisrow{_step}+200, y=AIME25, col sep=comma] {Figure/raw_data/stage2.csv};

      \node at (axis cs:\StageTwoMaxX,\StageTwoMaxY) [scale=0.7, star, star points=5,
      star point ratio=1.8, minimum size=2pt,
      fill=red!70!black, draw=red!70!black] {};
\node[anchor=west, text=red!70!black, font=\small]
      at (axis cs:\StageTwoMaxX-50,\StageTwoMaxY)
      {\pgfmathprintnumber[fixed,precision=1]{\StageTwoMaxY}\%};

      \node[anchor=south, font=\bfseries\small, text=gray!80]
  at (axis cs:100,23.7) {\textbf{\makecell{Exploration \\ (Parallel Thinking)}}};
  \node[anchor=south, font=\bfseries\small, text=gray!80]
  at (axis cs:375,23.7) {\textbf{\makecell{Exploitation  \\(Parallel $\rightarrow$ Sequential)}}};
      \addplot [very thick, dashed, gray!60] coordinates {(200,0) (200,27)};

    \end{axis}
    
    \begin{axis}[
        name=par,
        at={(acc.south west)}, anchor=south west,
        width=0.95\linewidth, height=8cm,
        xmin=0, xmax=550,
        ymin=0, ymax=100,
        ytick distance=20,
        axis y line*=right,
        axis x line=none,
        ylabel={\textbf{Parallel Ratio (\%)}},
        yticklabel style={color=ParLine, font=\small},
        y label style={color=ParLine, at={(axis description cs:1.15,0.3)}, anchor=west},
        every outer y axis line/.append style={ParLine},
        ytick style={ParLine},
        unbounded coords=discard
    ]

      \addplot+[color=ParLine, dashed, ultra thick, line width=1pt, mark=none,
                restrict x to domain=0:200]
        table[x=_step, y=ParallelRatio, col sep=comma] {Figure/raw_data/stage1.csv};
      \addplot+[only marks, mark=*, mark size=1.5pt,
                mark options={fill=ParLine, draw=white, line width=0.4pt},
                mark repeat=15, restrict x to domain=0:200]
        table[x=_step, y=ParallelRatio, col sep=comma] {Figure/raw_data/stage1.csv};

      \addplot+[color=ParLine, dashed, ultra thick, line width=1pt, mark=none,
                restrict x to domain=200:550]
        table[x expr=\thisrow{_step}+200, y=ParallelRatio, col sep=comma] {Figure/raw_data/stage2.csv};
      \addplot+[only marks, mark=*, mark size=1.5pt,
                mark options={fill=ParLine, draw=white, line width=0.4pt},
                mark repeat=15, restrict x to domain=200:550]
        table[x expr=\thisrow{_step}+200, y=ParallelRatio, col sep=comma] {Figure/raw_data/stage2.csv};

    \end{axis}

  \end{tikzpicture}
  \caption{Two-stage training with parallel reasoning as a mid-training exploration scaffold. Left axis plots AIME25 accuracy for Baseline (gray), Stage-1 (blue), and Stage-2 (red); right axis shows the proportion of outputs using the explicit parallel thinking structure. Stage-1 (0–200 steps; vertical dashed line) alternates ACC/PAR rewards to promote exploration, while Stage-2 continues GRPO with an accuracy reward only and is plotted after a +200-step shift to align the timeline. As training transitions from parallel to more sequential reasoning, the parallel ratio decreases yet accuracy continues to improve, peaking at 25.6\%, which exceeds single-thread model trained via GRPO.}
  \label{fig:mid_training}
\end{figure}

To empirically validate this hypothesis, we designed a two-stage training curriculum, with its dynamics and results presented in Figure \ref{fig:mid_training}.
\begin{itemize}[itemsep=0.0pt,topsep=0pt,leftmargin=*]
    \item \textbf{Stage-1 (Exploration Phase, steps 0-200)}: The primary goal of this initial phase is to maximize exploration. In this stage, we follow the training approach of our Parallel-R1-Unseen (S2), which explicitly incentivizes the use of the parallel thinking structure by applying an alternating ACC/PAR reward. As shown by the green dashed line in Figure \ref{fig:mid_training}, this successfully maintains a high parallel ratio, forcing the model to explore a wide breadth of reasoning paths constantly.
    \item \textbf{Stage-2 (Exploitation Phase, after 200 steps)}: At the 200-step mark, we change the focus from exploration to exploitation. The training objective is then switched to optimize for accuracy alone, allowing the model to refine and exploit the effective strategies discovered during the exploration phase.
\end{itemize}

The experimental results provide evidence in support of our hypothesis. As depicted in Figure \ref{fig:mid_training}, upon entering stage 2, the model's performance (red line) improves, reaching a peak AIME25 accuracy of 25.6\%, a notable improvement over the Baseline GRPO model. Critically, this performance gain occurs even as the model's reliance on the parallel structure decreases (as shown by the declining parallel ratio in stage 2). This key observation suggests that the value of parallel thinking lies not only in the effectiveness of the parallel structure itself (which already outperforms the baseline), but more importantly, in the robust policy space it helps discover through exploration. The initial forced exploration acted as a scaffold, guiding the model to a more effective region in the policy space, from which it could then learn a final policy.

\section{Conclusion}

In this work, we presented \textbf{Parallel-R1}, the \textbf{first reinforcement learning framework} to teach large language models to perform parallel thinking from scratch on real-world mathematical reasoning tasks. We proposed a \textbf{progressive training curriculum}, enabled by a simple and scalable data pipeline, that successfully bootstraps this complex skill by separating the learning of format, behavior, and core reasoning into distinct stages. Our approach achieved consistent accuracy improvements on several challenging mathematical reasoning benchmarks compared to strong baselines.

Our analysis yielded several key insights into the learning dynamics. We discovered that the model learns a risk-averse strategy, shifting its use of parallel thinking from early-stage \textbf{computational exploration} to late-stage \textbf{multi-perspective verification}. Most significantly, we empirically identified and validated the potential of parallel thinking as a \textbf{mid-training scaffold}, showing that adding this temporary, forced-exploration phase can unlock higher final performance ceilings after RL training.

\section*{Acknowledgment}
We thank Louis Liu from the University of California, Berkeley for insightful discussions and valuable feedback on this work.

\bibliography{ref}
\bibliographystyle{colm2024_conference}
\newpage
\appendix
\section{Prompts}

\begin{promptbox}[Baseline Prompt]
{Problem} Let's think step by step and output the final answer after "Final Answer: ".
\end{promptbox}

\begin{promptbox}[Parallel Thinking Prompt]
Solve the following problem step by step.
During the reasoning process, whenever you encounter a step that may benefit from multiple perspectives or independent reasoning, insert a <Parallel> block at that point.

Within each <Parallel> block:

Include at least two distinct and independent reasoning paths.
Each path must be enclosed within <Path> and </Path> tags.
Do not include any ordering information or cross-references between paths, as they are generated simultaneously and independently.
Close the block with </Parallel>.
Immediately after each </Parallel>, write a concise summary of insights or conclusions drawn from all paths, enclosed in <Summary> and </Summary> tags.

Repeat this process adaptively as needed throughout the reasoning.
Do not explicitly mention that you are triggering parallel thinking-just insert the <Parallel> block naturally within the reasoning chain.

End your response with a line starting with Final Answer: followed by the final result.

Problem: {Problem}
\end{promptbox}

\begin{algorithm}[H]
\caption{Parallel Thinking Format Check}
\label{alg:format-check}
\textbf{Input:} tokens – list of tokens from the parallel‑thinking trace; \\
\hspace*{2.1em} tag\_pairs – set of valid (opening, closing) tag pairs, e.g.\ \{(\verb|<Path>…</Path>|), …\} \\
\textbf{Output:} format\_valid – boolean indicating whether the trace is well‑formed
\begin{algorithmic}[1]
  \State $S \gets \emptyset$ 
  \State $\mathit{format\_valid} \gets \text{true}$
  \ForAll{$t$ \textbf{in} tokens}
    \If{$t$ \textbf{is an opening tag}}
      \State push $t$ onto $S$
    \ElsIf{$t$ \textbf{is a closing tag}}
      \If{$S$ is empty}
        \State $\mathit{format\_valid}\gets\text{false}$ 
        \State \textbf{break}
      \EndIf
      \State $\mathit{top\_tag}\gets\text{Top}(S)$
      \If{$(\mathit{top\_tag},\,t)\in\text{tag\_pairs}$}
        \State pop $S$
      \Else
        \State $\mathit{format\_valid}\gets\text{false}$
        \State \textbf{break}
      \EndIf
    \EndIf
  \EndFor
  \If{$\mathit{format\_valid}$ \textbf{and} $S\neq\emptyset$}
    \State $\mathit{format\_valid}\gets\text{false}$
  \EndIf
  \State \Return $\mathit{format\_valid}$
\end{algorithmic}
\end{algorithm}
 
\section{Case Studies}

\begin{figure}
    \centering
    \begin{tcolorbox}[
        colback=white,
        colframe=black!75,
        sharp corners,
        enhanced jigsaw,
        breakable,
    fontupper=\small
    ]
    
    \textbf{Case Study:} Find the sum of all integer bases $b>9$ for which $17_b$ is a divisor of $97_b$. \\[3pt]
    
    First, the number $17_b$ in base $b$ is equal to $1 \times b + 7 = b + 7$ in base 10. 
    Similarly, $97_b$ is equal to $9 \times b + 7 = 9b + 7$ in base 10. 
    We need to find all integer bases $b > 9$ such that $b + 7$ divides $9b + 7$.
    
    At this point, consider two distinct strategies for finding $b$:\\
    \begin{lstlisting}[basicstyle=\ttfamily\small, breaklines=true, breakatwhitespace=true]
<Parallel>
<Path>
Factor theorem approach: Consider 9b + 7 as a polynomial in b and use the factor theorem. 
Since b + 7 divides 9b + 7, by the factor theorem, 9(-7) + 7 = -63 + 7 = -56 must be divisible by b + 7. 
Since b > 9, b + 7 > 16, and -56 can be factored as -56x1, -28x2, -14x4, or -7x8 (and their positive counterparts). 
Only b + 7 = 14 gives b = 7, but b > 9, so no solution from this path. 
However, we should also consider the positive factors for completeness. 
The positive factorization 56x1, 28x2, 14x4, or 7x8 yields b + 7 = 56 (giving b = 49), 
b + 7 = 28 (giving b = 21), b + 7 = 14 (giving b = 7, invalid), 
and b + 7 = 8 (giving b = 1, invalid). 
Thus, the only valid solution from this path is b = 21 and b = 49.
</Path>
<Path>
Using base arithmetic: In base b, 97\_b - 1x17\_b = 97\_b - 17\_b = 80\_b, 
which is 8xb + 0 = 8b in base 10. 
Since 9b + 7 = 8b + (b + 7), b + 7 must divide 8b, implying that for some integer k, 
9b + 7 = (b + 7)(8 + k). 
Simplifying, 9b + 7 = 8b + 8k + kb + 7k, so b(1 - k) = 8k + 7. 
Since b > 9, 1 - k must be positive or negative to keep b an integer. 
If 1 - k = 1, then b = 8k + 7 with k < 1. 
Possible integer k values are k <= 0, but k = 0 gives b = 7 <= 9, 
and k < 0 gives negative b, which is invalid. 
Thus, this path leads to no valid solutions directly. 
However, it reinforces that b + 7 must divide 8b.
</Path>
</Parallel>
<Summary>
These two approaches consider both the algebraic factorization method 
and an alternative numerical or arithmetic approach to find possible values of b, 
but only the factorization method yields valid bases greater than 9.
</Summary>
\end{lstlisting}
Next, consider the condition that $b + 7$ divides $9b + 7$. 
From the earlier, $9b + 7 = 9(b+7) - 56$. 
For $b + 7$ to divide $9b + 7$, it must divide $56$. 
Since $b + 7 \ge 17$ for $b > 9$, the only divisors of 56 exceeding 16 are 28 and 56, giving $b+7\in\{28,56\}$ and thus $b\in\{21,49\}$.

Finally, verify:

For $b = 21$: $17_{21} = 21 + 7 = 28$, $97_{21} = 9 \times 21 + 7 = 196$, $196 \div 28 = 7$.\\
For $b = 49$: $17_{49} = 49 + 7 = 56$, $97_{49} = 9 \times 49 + 7 = 448$, $448 \div 56 = 8$.\\[4pt] 
Thus, the sum is $21 + 49 = 70$.    
Final Answer: 70
    \end{tcolorbox}
  \vspace{-0.1in}
  \caption{A qualitative case study showcases that our model uses parallel exploration of two distinct algebraic methods for solving a number theory problem, both converging on the correct answer.}
  \label{fig:case_1}
  \vspace{-0.1in}
\end{figure}

\begin{figure}
    \centering
    \begin{tcolorbox}[
        colback=white,
        colframe=black!75,
        sharp corners,
        enhanced jigsaw,
        breakable,
        fontupper=\small
    ]
    
    \textbf{Case Study:} Find the sum of all integer bases $b>9$ for which $17_b$ is a divisor of $97_b$. 

    First, I need to find all integer bases $b > 9$ for which $17_b$ is a divisor of $97_b$. To do this, I'll start by converting these base-$b$ numbers to base 10, as that's more familiar.

    In base $b$, the number $17_b$ means $1 \times b + 7 = b + 7$ in base 10. Similarly, $97_b$ means $9 \times b + 7 = 9b + 7$ in base 10.

    So, the condition that $17_b$ divides $97_b$ translates to $b + 7$ dividing $9b + 7$ in base 10.

    Now, for $b + 7$ to divide $9b + 7$, the remainder when $9b + 7$ is divided by $b + 7$ must be zero. I can perform polynomial division or use the remainder theorem to find this.

    Let me compute the remainder of $9b + 7$ divided by $b + 7$.
    
    Divide $9b + 7$ by $b + 7$:
    - The quotient is 9, since $9 \times (b + 7) = 9b + 63$.
    - Subtract this from $9b + 7$: $(9b + 7) - (9b + 63) = 9b + 7 - 9b - 63 = -56$.

    So, the remainder is $-56$. For $b + 7$ to divide $9b + 7$, it must divide $-56$, or equivalently, $b + 7$ must be a positive divisor of 56.

    The positive divisors of 56 are: 1, 2, 4, 7, 8, 14, 28, 56.
    
    Since $b > 9$, we have $b + 7 > 16$. So, $b + 7$ must be a divisor greater than 16.
    - $b + 7 = 28 \Rightarrow b = 21$ (valid)
    - $b + 7 = 56 \Rightarrow b = 49$ (valid)

    So, the valid bases are $b = 21$ and $b = 49$. Now, I need to find the sum: $21 + 49 = 70$.

    But before finalizing, I should confirm that these are the only valid bases. From the divisor check, only $b + 7 = 28$ and $b + 7 = 56$ give $b > 9$, so no others. 
\begin{lstlisting}[basicstyle=\ttfamily\small, breaklines=true, breakatwhitespace=true]
<Parallel>
<Path>One might consider that $ b + 7 $ not only divides 56 but also that it must be such that $ 17_b $ and $ 97_b $ are valid base-$b$ numbers, which requires $ b > 7 $ (since digits 9 and 7 are used). However, since $ b > 9 $ is given, this is already satisfied, and the divisor condition is the key.</Path>
<Path>Use the fact that for $ b + 7 $ to divide $ 9b + 7 $, the remainder must be zero, so we can solve $ 9b + 7 = k(b + 7) $ for integer $ k $, but it's simpler to compute the remainder directly as above.</Path>
</Parallel>
<Summary>
The parallel block explored alternative strategies for solving $ b + 7 $ divides $ 9b + 7 $, including finding divisors, considering base validity, and solving equations, but ultimately confirmed that only $ b = 21 $ and $ b = 49 $ satisfy the condition for $ b > 9 $.
</Summary>
\end{lstlisting}
Now, sum the valid bases: $21 + 49 = 70$.
Final Answer: 70
\end{tcolorbox}
  \vspace{-0.1in}
 \caption{
  A qualitative case study demonstrates late-stage verification through parallel thinking.}
  \label{fig:case_2}
  \vspace{-0.1in}
\end{figure}

\end{document}